\documentclass[letterpaper, 10 pt, conference]{ieeeconf}
\IEEEoverridecommandlockouts
\usepackage{cite}
\usepackage{amsmath,amssymb,amsfonts}
\usepackage{algorithmic}
\usepackage{graphicx}
\usepackage{textcomp}
\usepackage{xcolor}
\usepackage{subcaption}
\usepackage{url}

\def\BibTeX{{\rm B\kern-.05em{\sc i\kern-.025em b}\kern-.08em
    T\kern-.1667em\lower.7ex\hbox{E}\kern-.125emX}}
\begin{document}

\title{Leveraging Augmented Reality for Improved Situational Awareness During UAV-Driven Search and Rescue Missions}

% \author{Anonymous for Double Blind Review}
\author{
    Rushikesh Nalamothu, Puneet Sontha, Janardhan Karravula, and Ankit Agrawal\\
    Department of Computer Science, Saint Louis University, Saint Louis, MO\\
    \{rushikesh.nalamothu, puneet.sontha, janardhan.karravula, ankit.agrawal.1\}@slu.edu
    }
    
\maketitle

\begin{abstract}
In the high-stakes domain of search-and-rescue missions, the deployment of Unmanned Aerial Vehicles (UAVs) has become increasingly pivotal. These missions require seamless, real-time communication among diverse roles within response teams, particularly between Remote Operators (ROs) and On-Site Operators (OSOs). Traditionally, ROs and OSOs have relied on radio communication to exchange critical information, such as the geolocation of victims, hazardous areas, and points of interest. However, radio communication lacks information visualization, suffers from noise, and requires mental effort to interpret information, leading to miscommunications and misunderstandings. To address these challenges, this paper presents VizCom-AR, an Augmented Reality system designed to facilitate visual communication between ROs and OSOs and their situational awareness during UAV-driven search-and-rescue missions. Our experiments, focus group sessions with police officers, and field study showed that VizCom-AR enhances spatial awareness of both ROs and OSOs, facilitate geolocation information exchange, and effectively complement existing communication tools in UAV-driven emergency response missions. Overall, VizCom-AR offers a fundamental framework for designing Augmented Reality systems for large scale UAV-driven rescue missions.

% This paper explores the integration of Augmented Reality(AR) with Uncrewed Aerial Vehicles(UAVs) in emergency response missions. It shows the potential of AR in improving communication between Remote Operators(ROs) and On-Site Operators(OSOs), and enhancing the spatial awareness of ROs and OSOs. The current practice involves ROs monitoring UAV aerial stream and communicating crucial information to OSOs using radio which often leads to miscommunication. The paper highlights the challenges of radio communication in conveying geolocation details and proposes VizCom-AR, an AR system designed to address the communication challenges by enabling the visualization of geolocation information. VizCom-AR enables ROs to mark Points of Interest (POIs) on UAV aerial streams, while OSOs can monitor these POIs on the scene in real time through Microsoft HoloLens AR glasses. To validate the proposed system, a user study is conducted and the results indicate marked improvements over radio communication. Overall, VizCom-AR contributes to the synergistic
% integration of AR and UAV technologies to support emergency response operations personnel, laying the foundation for further
% interdisciplinary work.
\end{abstract}

% \begin{IEEEkeywords}
%  Uncrewed Aerial Vehicles, Augmented Reality, Geo-spatial Intelligence
% \end{IEEEkeywords}
\newcommand{\dronear}{RO-AR }
\newcommand{\server}{Server }

\newcommand{\Dronear}{RO-AR}
\newcommand{\dronearx}{RO-AR} %nospace

\newcommand{\humanar}{OSO-AR }
\newcommand{\Humanar}{OSO-AR}

\newcommand{\missionplan}{Mission-Control }
\newcommand{\Missionplan}{Mission-Control}

\newcommand{\system}{VizCom-AR }
\newcommand{\systemx}{VizCom-AR}
\newcommand{\System}{VizCom-AR}

\section{Introduction}

%%%%% General to Specific Intro
    Unmanned Aerial Vehicles (UAVs) are transforming public safety operations such as search and rescue \cite{khan2014information}, firefighting \cite{jayapandian2019cloud}, and disaster relief, by providing an aerial perspective of the scene. For instance, law enforcement officers in California utilized smart UAVs to track a shooter, relying on live video feeds monitored by off-site officers. Similarly, during the Notre-Dame cathedral fire in Paris in 2019, UAVs equipped with thermal cameras were deployed in coordination with over 400 rescue team members \cite{DJIdrone93:online}. The sheer size of the response highlights the need for effective communication between humans, smart UAVs, and between humans and UAVs during emergency situations \cite{cleland2020human}.

% \footnote{\url{https://bit.ly/drones-in-firefighting}}

During UAV-assisted public safety operations, specific team members, known as Remote Operators (ROs), analyze aerial video footage from the UAVs. This footage offers a bird's-eye view of the scene, enhancing the ROs' situational awareness and enabling them to identify and examine critical details of the environment. The ROs relay their insights to the On-Site Operators (OSOs), who are actively engaged in ground activities at the physical scene. In current practice, ROs and OSOs communicate via radio, exchanging vital \emph{Scene Information} including  geocoordinates of individuals in need of assistance, location of potential evidence that needs further investigation, and areas that might pose hazards for OSOs. Additionally, ROs provide planning advice to OSOs, guiding them on prioritizing targets to accomplish mission objectives efficiently and safely. OSOs also communicate their mission progress back to ROs.

\begin{figure}[t]
    \centering
    \captionsetup{justification=centering}
    \includegraphics[width=0.97\columnwidth]{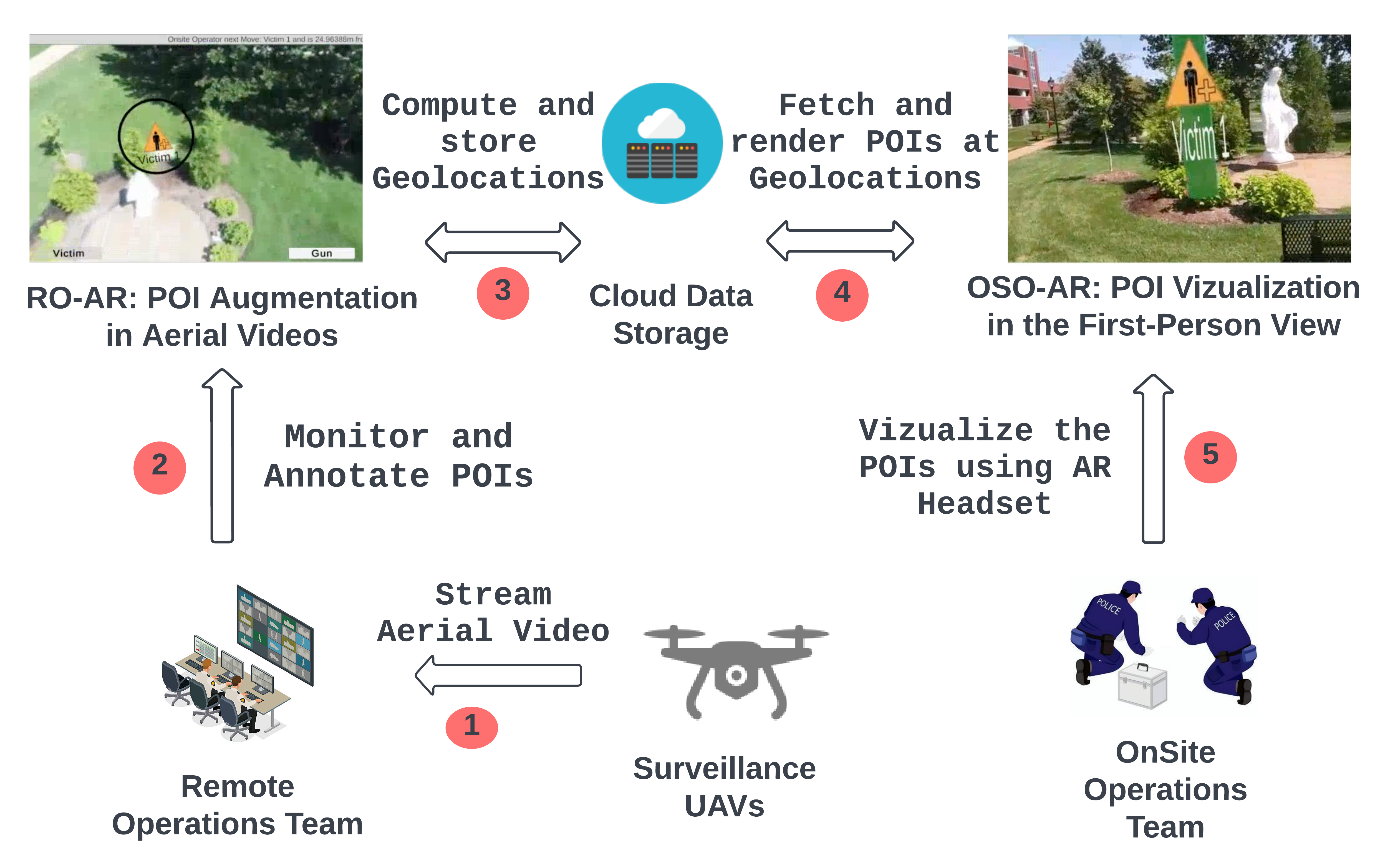}
    \caption{\system: Visual Communication Overview}
    \label{fig:overview}
\end{figure}
The current public safety reliance upon radios, to relay essential scene information, presents unique challenges, particularly when communicating geolocation details such as the exact coordinates of a victim, evidence spots, or a suspect's last known location. In our prior work \cite{agrawal2020next}, we conducted design sessions with Fire-fighters and found that sharing information in noisy environments via radio often results in miscommunication and misunderstandings, and therefore decreased operators' situational awareness. In particular, we found that conveying geo-coordinates verbally introduced ambiguities due to the absence of visual context, interference on radio channels, and the cognitive load involved in accurately processing multiple sets of coordinates. During time-sensitive, critical situations, this lead to confusion, delays, and, loss of situational awareness. Moreover, radio communication makes it difficult to ensure consistent interpretation of the scene in large-scale rescue missions involving hundreds of team members \cite{manoj2007communication}. These issues  highlight the need for designing innovative systems that enable rescue team members to share and visualize precise geolocatable information during rescue and emergency response missions.

With the advancement of UAVs, Augmented Reality (AR) technology has also evolved, offering visualization support by overlaying digital information onto the real world. While AR has gained popularity in fields such as healthcare and education, its application in UAV-driven search and rescue missions remains under-explored. Addressing this research gap requires a collaborative effort across academia, industry, and rescue personnel to design effective, useful, and innovative solutions. Therefore, in this paper, we present our AR platform called \System.

We utilize the paradigms of Location-based Augmented Reality to design \system to help ROs and OSOs communicate scene information and maintain situational awareness. The design of \system comprises two different AR applications: \humanar and  \dronear that together empower responders to share and visualize scene information according to the diverse needs of their specific roles. Figure \ref{fig:overview} provides an overview of our system during a UAV-driven search-and-rescue operation. ROs use \dronear to annotate the live video stream with Points of Interest (POIs), and OSOs use \humanar to vizualize the same POIs in their first-person view using AR-headset.  We conducted multiple studies to learn the feasibility, usability, and end-users' (police officers) perspective of our system. We found that \system makes critical geolocation information more comprehensible, and that visualization of the geolocation improves the spatial understanding of the scene among the rescue team members. Further, a team of police officers indicated that \system could effectively complement the existing radio communication infrastructure of their emergency rescue teams.

% \begin{itemize}
%     \item First, we employ the fundamental principles of the pin-hole camera model to compute geolocation of POIs in UAV aerial video streams. we evaluate this approach under a variety of UAV operational scenarios, including variations in altitude, and camera orientations (pitch).

%     \item Second, we discuss the overall design of our augmented reality system for UAVs. Additionally, we carry out an assessment of the precision with which these markers are visualized over the aerial video streams.
% \end{itemize}
%%%%%%%%%%%%%%% Paper Structure
% The remainder of this paper is organized as follows. Section \ref{sec:related-work} reviews related work at the intersection of Augmented Reality and robotics, and explores innovative applications of AR in emergency response. Section \ref{sec:ar_system_design} discusses our UAV-centric approach as well as the design of \System. Section \ref{sec:exp} discusses a the performance of \system and compare it with Radio-based communication systems. \ref{sec:discussion} briefly discusses the wider applicability of our approach in various emergency response as well as its scalability for large-scale multi-UAV missions. Finally, Section \ref{sec:conclusion} concludes our study. 
\section{Related Work}
\label{sec:related-work}
% Agrawal et al.,\cite{cleland2020requirements,agrawal2020model,agrawal2020next} conducted a series of participatory design sessions with emergency service providers to explore the role that smart UAVs could play in assisting emergency situations. They also provided guidance for designing interfaces for UAV-based emergency response systems and emphasized the need to support human collaboration \cite{cleland2020human, abraham2021adaptive}. While prior research explored the significance of humans collaborating with multiple UAVs under various circumstances to achieve mission goals; the research described in this paper, presents a novel way to improve communication quality among human members of a mission team through annotating aerial video streams with Points of Interest (POI) information and visualizing it across role-specific interfaces. Further, our design of \system framework aligns with the vision of Lalone et al., \cite{lalone2019vision} and Alharti et al., \cite{alharthi2021activity}, which recommend drawing ideas from mixed reality games, and AR technology to design future collaborative information systems for emergency response. 

\subsection{Augmenting Reality Systems for Robots}
The comprehensive survey conducted by  Suzuki et al., \cite{suzuki2022augmented}, on AR-enhanced Robotic Interfaces, highlights various applications in which AR significantly enhances human-robot collaboration. For instance, Boateng et al.,\cite{boateng2023iros} leveraged AR to enhance Team Situation Awareness in human-robot interactions. In the context of UAVs, Erat et al.,\cite{erat2018drone}, used AR to simplify UAV path planning in spatially restricted settings by offering an exocentric viewpoint to operators, and Liu et al.,\cite{liu2020iros} developed an AR system for controlling an autonomous UAV using hand movements on a 3d Map augmented inside a Hololens. 

Broadly, current AR solutions in the domain of UAV-driven missions primarily leverage AR to (a) vizualize UAV flight paths \cite{li2015flying,walker2018communicating,paterson2019improving} (b) control or send commands to one or more UAVs \cite{hedayati2018improving,stevenson2015beyond,pittman2014exploring,erat2018drone}, and/or to (c) augment aerial video streams with previously known location-based information such as landmarks \cite{crowley2014ar,sun2013real}. In contrast, one of the core contribution in this domain is our method for augmenting aerial videos in real-time with the new or previously unknown information discovered by operators during the mission and share it with other mission stakeholders. 

% \textbf{Aerial Video Augmentation Systems:}
% AerialAR \cite{crowley2014ar} accesses data from map services such as Google Maps \cite{GoogleMa74:online} to obtain geo-information about the area visible through the UAV's camera and then overlays it on aerial video streams. Sun et al.,\cite{sun2013real}, proposed a system that overlays the height information of each building visible in the aerial video stream. Similarly, Unal et.al,\cite{unal2020distant}, leveraged location-based approaches to augment the aerial video stream with predefined 3D models. These systems overlay the aerial video stream with geo-referenced information that is known in advance and remains independent of the emergency situation. Furthermore, existing AR-based solutions do not allow humans to dynamically create AR content during a mission. In contrast, our system allows augmentation of the aerial video stream with mission-specific information, such as the location of victims, last known locations of a suspect, or location of evidence found on the scene, and also allows ROs to create AR content dynamically at run-time to support OSOs' in navigation and improving their situational awareness. 

\subsection{AR Systems in Emergency Response}
The THEMIS-AR \cite{nunes2018augmented} is an AR mobile app designed to improve OSO's scene perception by overlaying context-relevant information, such as distance, time, and position of  POIs on their current view of the scene via a mobile device during a standard non-UAV based emergency response mission. Further, Campos et al., \cite{campos2019mobile} extended THEMIS-AR to study AR systems and provided guidelines for designing AR applications for emergency response. However, the scene information in THEMIS-AR is geo-referenced manually by RO and the AR application overlays that same information on the mobile device of OSO. Further, these systems do not allow ROs to share information back to OSOs via their AR mobile devices, thereby blocking bi-directional communication between OSOs and ROs. In contrast, our approach allows OSOs to (a) visualize dynamically identified scene information by ROs in the first-person view via an AR-headset, and (b) share information back to RO about the scene, such as their progress or plans for completing the ground activities with ROs, establishing the bi-directional information exchange.

\section{\system}
\label{sec:ar_system_design}
% \subsection{System Overview}

The \system system consists of two AR interfaces: \dronear for ROs and \humanar for OSOs. ROs, monitoring UAV footage, use \dronear to mark POIs on the video, which are then geo-located and shared with OSOs. OSOs, equipped with AR headsets such as Microsoft Hololens, use \humanar to view those POIs, gain additional information such as distance to POIs, and communicate their status and plans to conduct ground operations such as which POI to investigate next. Figure \ref{fig:overview} illustrates the application of our system in real-world. Next, we discuss the primary components of both \dronear and \humanar systems.

\subsection{\dronear Design}
To design \Dronear, we needed to compute the geolocation of POIs visible in the aerial video and augment the live aerial video stream with an AR marker in real-time. 

\noindent\textbf{Geo-Location Computation}:
\label{subsec:geo-loc_computation}

\begin{figure}[htbp]
    \centering
    \includegraphics[width=0.9\columnwidth]{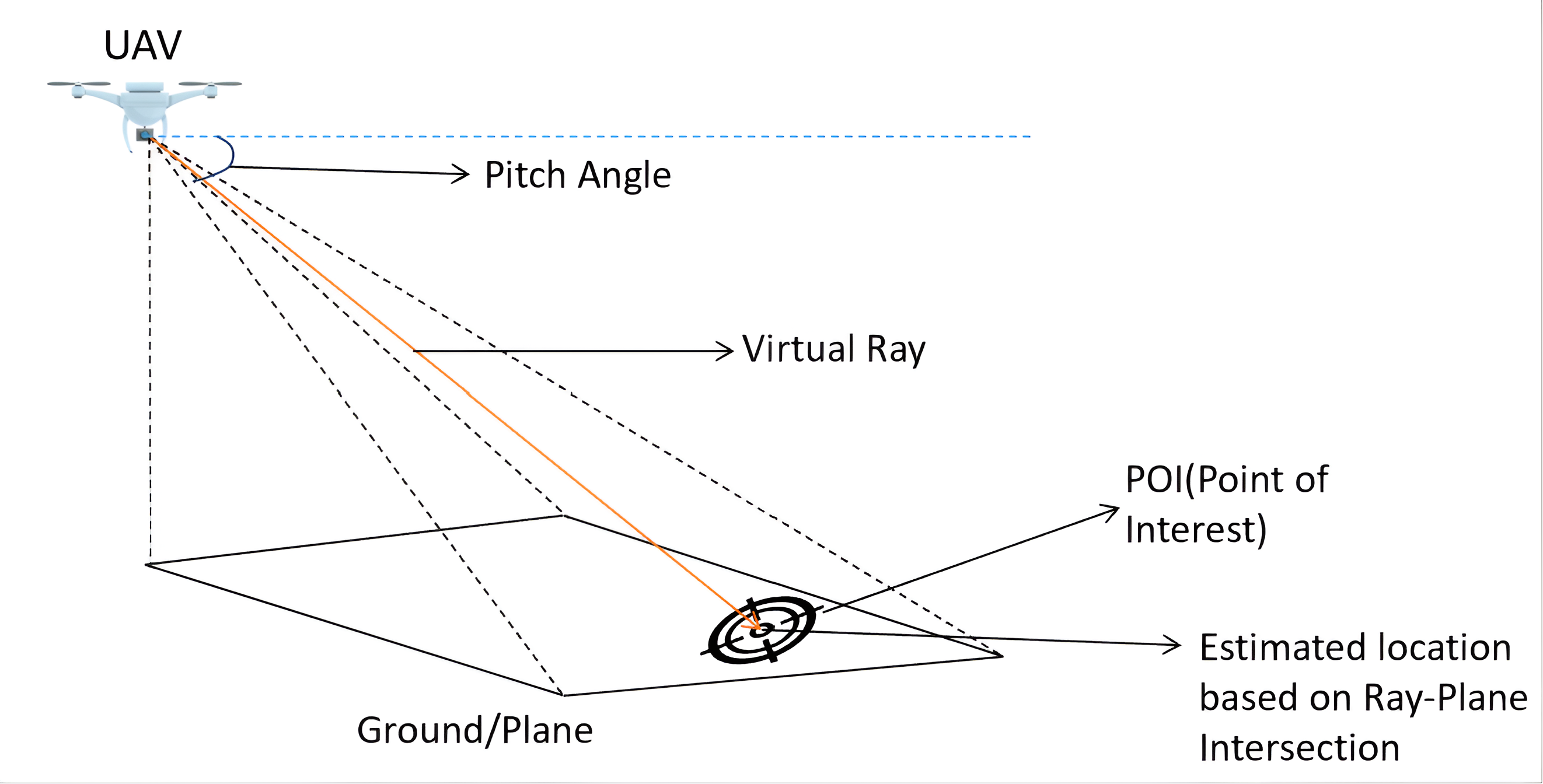}
    \caption{Geo-location computation}
    \label{fig:geo-compute}
\end{figure}
We employ the pinhole camera model to calculate the intersection of a UAV camera's line of sight with the Earth's surface. We treat the UAV's camera as a pinhole camera, with each image sensor pixel representing a unique line of sight in 3D space. When a POI is identified by RO and selected in the aerial video stream, we project a directed ray from the camera through the identified pixel. The intersection point of this ray with the Earth's surface determines the POI's geolocation. During our calculation, we utilize the standard WGS-84 to calculate geographical positions of POIs in terms of latitude, longitude, and elevation above the ellipsoid of Earth. Figure \ref{fig:geo-compute} illustrates the application of pinhole camera model to compute the geolocation of POI visible in aerial video frame.

\begin{figure}[htbp]
    \centering
    \includegraphics[width=0.9\columnwidth]{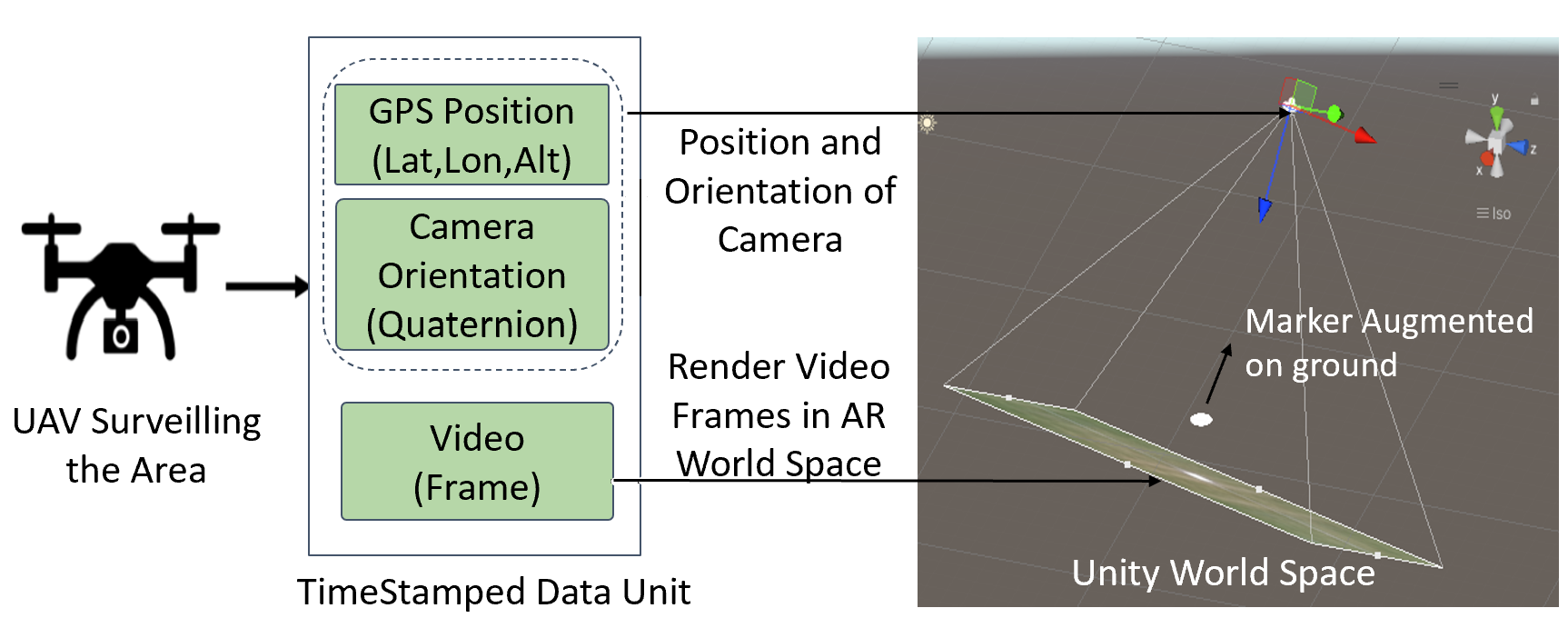}
    \caption{Marker augmentation in aerial video stream}
    \label{fig:flying-ar-implementation}
\end{figure}

% \begin{figure}
%     \centering
%     \includegraphics[width=\columnwidth]{figures/analysis/RO-AR-Interface.jpg}
%     \caption{RO-AR interface showing notifications, OSO live location, annotated POIs, OSO's next target, and distance of OSO from his next target }
%     \label{fig:ro-poi}
% \end{figure}

\noindent\textbf{AR Marker Augmentation in Aerial Video}
\label{subsec:marker_augmentation}
A critical aspect of \dronear is the real-time placement and augmentation of POI markers in the aerial video frame. These POI markers are positioned in the AR World Space of \dronear based on their geolocation data (computed as discussed before), and placed at ground level (zero meters altitude). The precision of UAV camera's movement in AR World Space depends on accurately interpreting data received from the UAV; GPS coordinates (latitude, longitude, altitude), and camera orientation data in Quaternions. GPS data from UAVs are converted into Unity's local coordinate system in real-time to position the virtual camera precisely in AR space, mimicking the UAV's real-world movements. The virtual camera's orientation is adjusted using Quaternion values to align with the UAV camera's perspective on the ground at each time step. Additionally, virtual camera properties such as lens focal length and field of view are configured to match those of the real UAV camera. As the AR camera moves and navigates through the AR environment guided by the UAV's movements these POI markers become visible in the video feed when they enter the virtual camera's field of view. The AR World Space illustrating this augmentation is shown in Figure \ref{fig:flying-ar-implementation}, which shows the AR camera's field of view rendering the aerial video frames as well as a POI marker (as overlay) within this view.

\begin{figure}[t]
    \centering
    \includegraphics[width=\columnwidth]{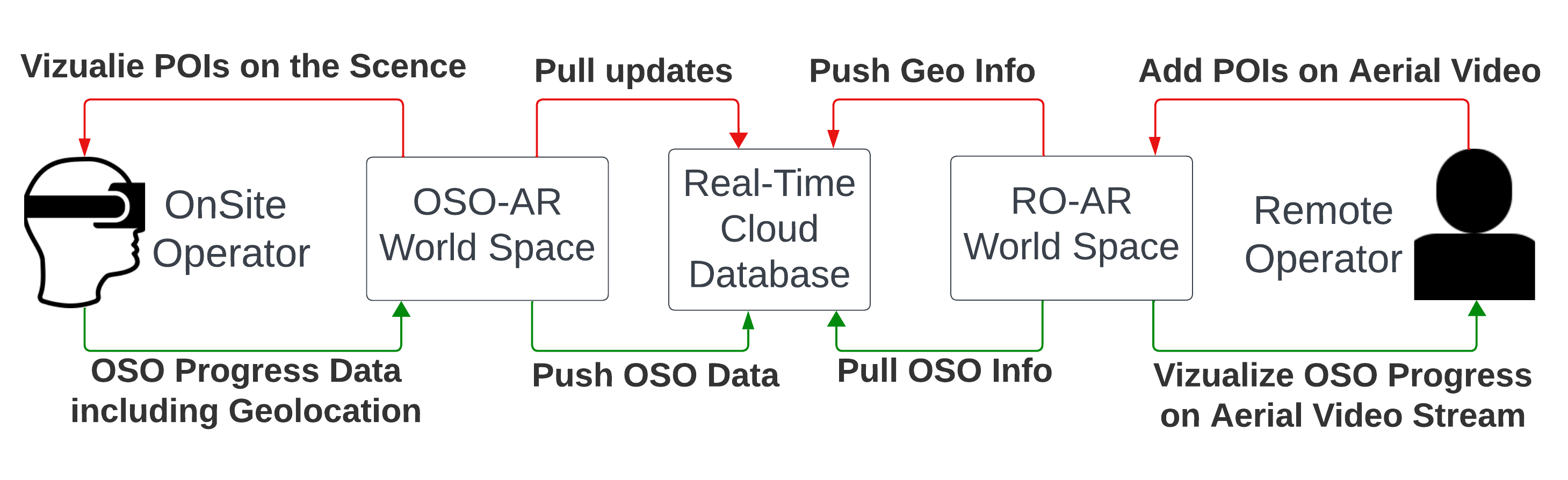}
    \caption{The bi-directional data flow between ROs and OSOs}
    \label{fig:firstresponse-ar-implementation}
\end{figure}
\begin{figure*}[htbp]
    \centering
    % \captionsetup{justification=centering}
    \begin{subfigure}[b]{0.32\textwidth}
        \centering
        \includegraphics[width=\textwidth]{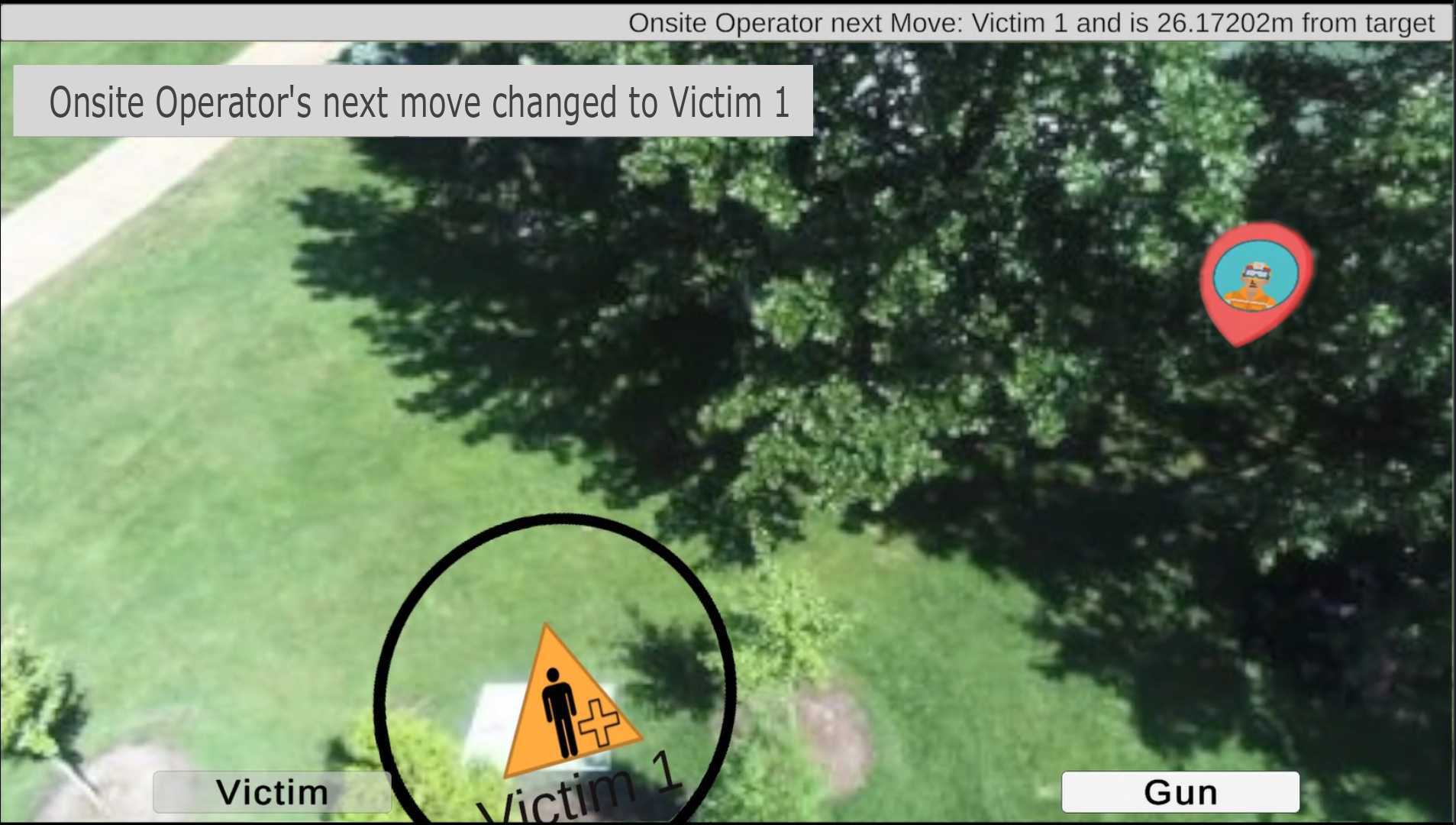}
        \caption{Live view of RO-AR during the mission.}
        \label{fig:ro-poi}
    \end{subfigure}
    \hfill
    \begin{subfigure}[b]{0.32\textwidth}
        \centering
        \includegraphics[width=\textwidth]{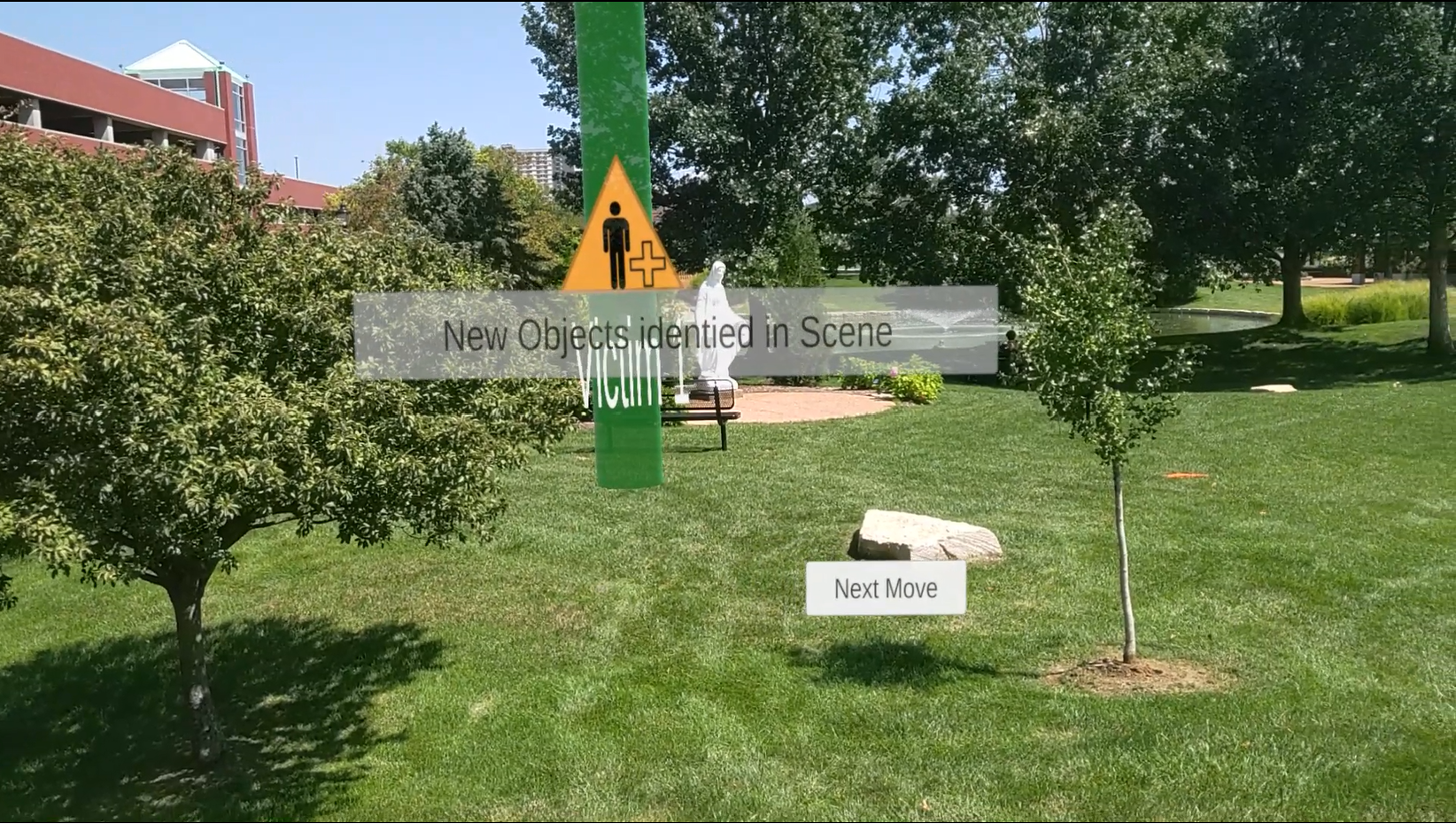}
        \caption{Live view of OSO during the mission }
        \label{fig:oso-poi}
    \end{subfigure}
    \hfill
    \begin{subfigure}[b]{0.32\textwidth}
        \centering
        \includegraphics[width=\textwidth]{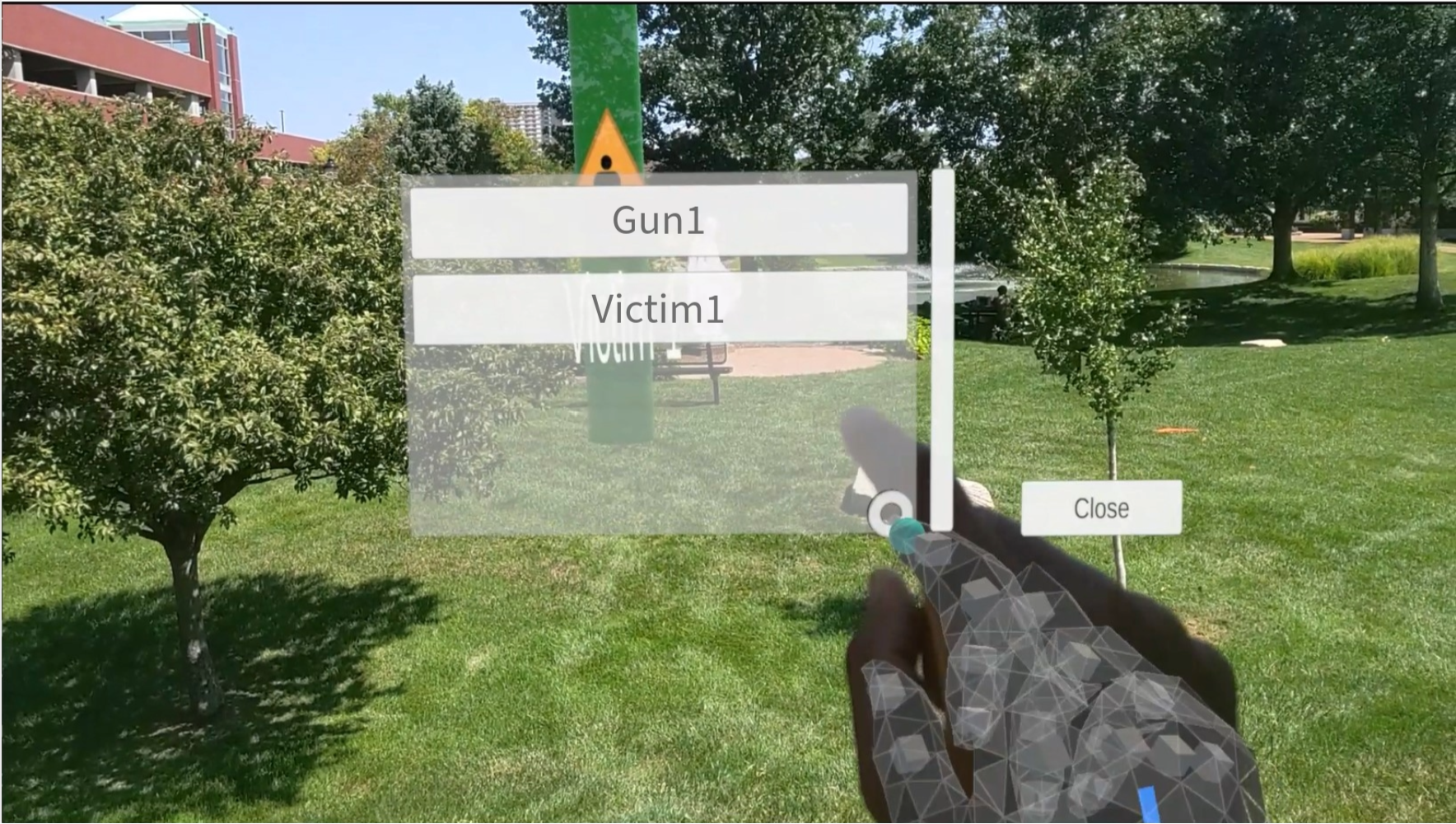}
        \caption{ OSO-AR panel to send field updates}
        \label{fig:oso-panel}
    \end{subfigure}
    \caption{RO-AR and OSO-AR Interfaces: (a) System displaying notifications - OSO live location, annotated POIs, OSO's next target, and the distance of OSO from their next target; (b) OSO utilizing AR glasses during ground operations; (c) OSO communicating the next search goal to RO on the ground.}

    \label{fig:combined-figures}
\end{figure*}

\subsection{\humanar}
\humanar is a Location-Based AR system for on-site officers or OSOs with two important goals. First, to establish a bi-directional visual communication channel between OSOs and ROs, enabling OSOs to share their mission progress with ROs. Second, to visualize geolocated Points of Interest using Microsoft HoloLens 2, supporting situational awareness and decision-making.

\noindent\textbf{Bidirectional Information Sharing:} \humanar is a Unity based application and creates its own AR world space where its AR camera replicates the real-world position and settings of the OSO's headset. POIs are added in \Humanar's world space through \system's cloud database over Wi-Fi. When an RO using \dronear adds a POI, the geolocation of the POI and its metadata such as type of POI are recorded in \systemx's database. To keep \Humanar's AR World updated with POIs, \humanar fetches data from this database every 5 seconds. Also, \humanar shares data like the OSO's location with \dronear through the database, helping the RO track the mission. Figure \ref{fig:firstresponse-ar-implementation} illustrates the bidirectional data flow between \humanar and \dronear.

\label{interface-dronear}
 
% The \humanar interface extends Campos et al., THEMIS-AR \cite{campos2019mobile} design for use in a multi-UAV domain. While THEMIS-AR requires humans to manually provide the geolocation of the POI in the form of GPS coordinates, our solution lifts this limitation by automatically computing the geolocations of POIs based on annotations in the aerial video stream via \dronearx.  
% Figure \ref{fig:firstresponse-ar-implementation} shows the \humanar application interface.\\ 

\noindent\textbf{POI Visualization:}
In the \humanar system, POIs are visualized as 5-meter-tall virtual cylinders in the AR space, with an icon and label (indicating the POI type, like 'victim' or 'evidence') positioned 2 meters above the ground as shown in Figure \ref{fig:oso-poi}. These cylinders automatically increase in height by 0.02 meters for every meter beyond 10 meters from the OSO, ensuring visibility in crowded or distant areas. The labels and icons are designed to rotate towards the OSO for improved visibility. For mission progress updates, \humanar includes an AR panel that allows OSOs to inform the RO about their mission status and select their next POI for investigation as shown in Figure \ref{fig:oso-panel}.

\section{Preliminary Discussion with Police Officers}

\label{sec:domain_experts}

We conducted a focus group session to gather feedback from the end-users of \system on the conceptual design. Invitations to the focus group were sent via emails and resulted in the participation of five police officers (four males and one female). The participating officers were highly experienced in investigating incidents, with experience ranging from 15 to 40 years. Table \ref{tab:participants_details} shows the participant's backgrounds. Participants of the focus group were asked to discuss openly their past experiences of using UAVs, and opinions about \System. 

\begin{table}[htbp]
    \centering
        
    \begin{tabular}{|c|c|c|} \hline
         \textbf{Id}& \textbf{Current Designation}&\textbf{Years of Service}\\ \hline
         
P1& Police Chief &17\\  \hline
P2&Deputy Chief of Police&27 \\ \hline
P3&Assistant Police Chief&40\\ \hline
P4&Special Events Program Manager&15\\ \hline
P5&Deputy Chief Safety Services&34\\ \hline

    \end{tabular}
\caption{Details of police officers in our focus group}
    \label{tab:participants_details}
\end{table}

In the first half of the session, we discussed UAV applications in emergency response, including river search and rescue, fire surveillance, and city surveillance, supported by real-world examples from news articles. We then explored participants' current communication practices and challenges during emergencies. In the second half of the session, we presented the design of \system using images and videos showcasing \humanar and \dronear in operation. We encouraged discussion to understand their perspectives on the system's potential applications and took notes for analysis.

\subsection{Insights Gained from Focus Group Session}
During the session, the police officers confirmed our previous understanding that they communicated primarily verbally via radio. Most participants agreed that radio communication suffers from excessive chatter and that superfluous information is frequently broadcast when multiple people talk or communicate simultaneously. Moreover, \textit{P5} pointed out that, in large-scale operations, this can create a chaotic situation, resulting in confusion among rescue personnel. \textit{P1} also stated that radio communications between dispatchers and callers can be frustrating due to the time-sensitive nature of the situation. 

While discussing how geolocation information is communicated over the radio and how police officers interpret it, we learned that officers frequently refer to landmarks, roads, buildings, and other known entities in their environment. In this context, \textit{P1} shared that it can be difficult to identify locations when several buildings have similar names or textures. Police Officers' current communication infrastructure lacked the capability to visualize information in any form. Finally, police officers reported that geolocation assists them in many ways -- for example, by narrowing down their search or containment areas. They also agreed that tracking down a suspect becomes easier if the suspect's past geolocations are known. Overall, during the discussion, police officers provided positive feedback on \system to alleviate communication challenges, validating the design of \System. 

\subsection{\system Evaluation Goals}
After receiving positive feedback from police officers, we proceeded with the system's development and conducted a formal analysis of its effectiveness during search and rescue operations. Our primary goal was to evaluate \system in real-world scenarios aimed at addressing the following research questions:

\noindent\textbf{RQ1:} How does the geolocation computation and marker augmentation of \dronear vary across diverse flight characteristics, such as altitude and camera pitch angles?

\noindent\textbf{RQ2:} How does the overall design of \system system compare to radio communication when communicating critical scene information?

% \subsection{Police Officer's Perception of our conceptual framework}

% We developed a short questionnaire to obtain feedback from police officers regarding their perception of the usability of the \system during crime investigations, and asked police officers to rate their agreement or disagreement with the statements in Table \ref{tab:survey_questions} on a 5-point Likert scale. The survey results suggest the positive feedback from the end-users of \system and validates the design of \system to improve the communication among police officers during emergency response.

% \input{tables/survey_results}
% \begin{figure*}[htbp]
% \begin{minipage}{.5\linewidth}
%   \input{tables/survey_results}
%   \end{minipage}%
%   \begin{minipage}{.5\linewidth}
%   \centering
%     \includegraphics[width=\columnwidth]{figures/analysis/PoliceND_Survey.png}
%     \caption{Officers' agreement with the survey questionnaire}
%     \label{fig:police_survey_questions}
%   \end{minipage}
% \end{figure*}
% \section{Experiments and Analysis}
% \label{sec:exp}

\section{RQ1 - Experiments and Analysis}
\subsection{Data Collection}
We performed multiple carefully planned drone flights to collect video data that captures a POI in the scene. We also carefully placed a small whiteboard in an open field, marking it as our main POI for evaluation purpose. We accurately recorded its geo-location and also cross-verified this location manually using Google Maps. This thorough approach ensured a robust ground truth for our study. 

Throughout the data collection process, we controlled the key variables of UAV to capture POI in the frame from various angles, heights, and distances. We collected aerial videos from three varying heights: 10 meters, 20 meters, and 30 meters; the latter being approximately 98 feet, which is the safe maximum allowed flight altitude in the area. We chose these heights to give us a range of views, from close up to a moderate distance above the ground. In addition to adjusting the height, we also changed the pitch angle of the UAV to four different angles: 45, 60, 75, and 90 degrees (Perpendicular to the ground). Each angle gave us a different way of looking at the POI. 

\subsection{Geo-location Computation Analysis Method}
In order to answer RQ1, we computed the (a) horizontal distance between the actual location of a POI in the real world and the computed location by our system and (b) the pixel distance between the AR marker and actual POI as visible in the aerial video stream. We utilized \dronear to compute the geolocation of the POI as seen in the collected videos. During the analysis process, our objective was to precisely click at the center of the POI visible in the video frames. Since the analysis based on manual clicks could be compromised by the precision of the click itself, we automated the process of clicking on the POI and recording its pixel coordinates. As a result, from each video, we selected ten frames in which the whiteboard (POI) was visible and located at different parts of the frame. For each frame, we applied a common image processing technique known as contour detection using openCV \cite{2015opencv} to identify the region of the screen occupied by POI and recorded the pixel coordinate of the POI's center on the frame. These pixel coordinates of POI and the frame metadata are then used to calculate the location of POI as discussed in Section \ref{subsec:geo-loc_computation}. The geolocation computed by this method is then compared with the actual ground truth geolocation of the POI.

\subsection{Analysis - Geo-location Computation Accuracy}  
\begin{figure}[t]
    \centering
    \captionsetup{justification=centering}
    
    \begin{minipage}{0.86\columnwidth}
        \includegraphics[width=\columnwidth]{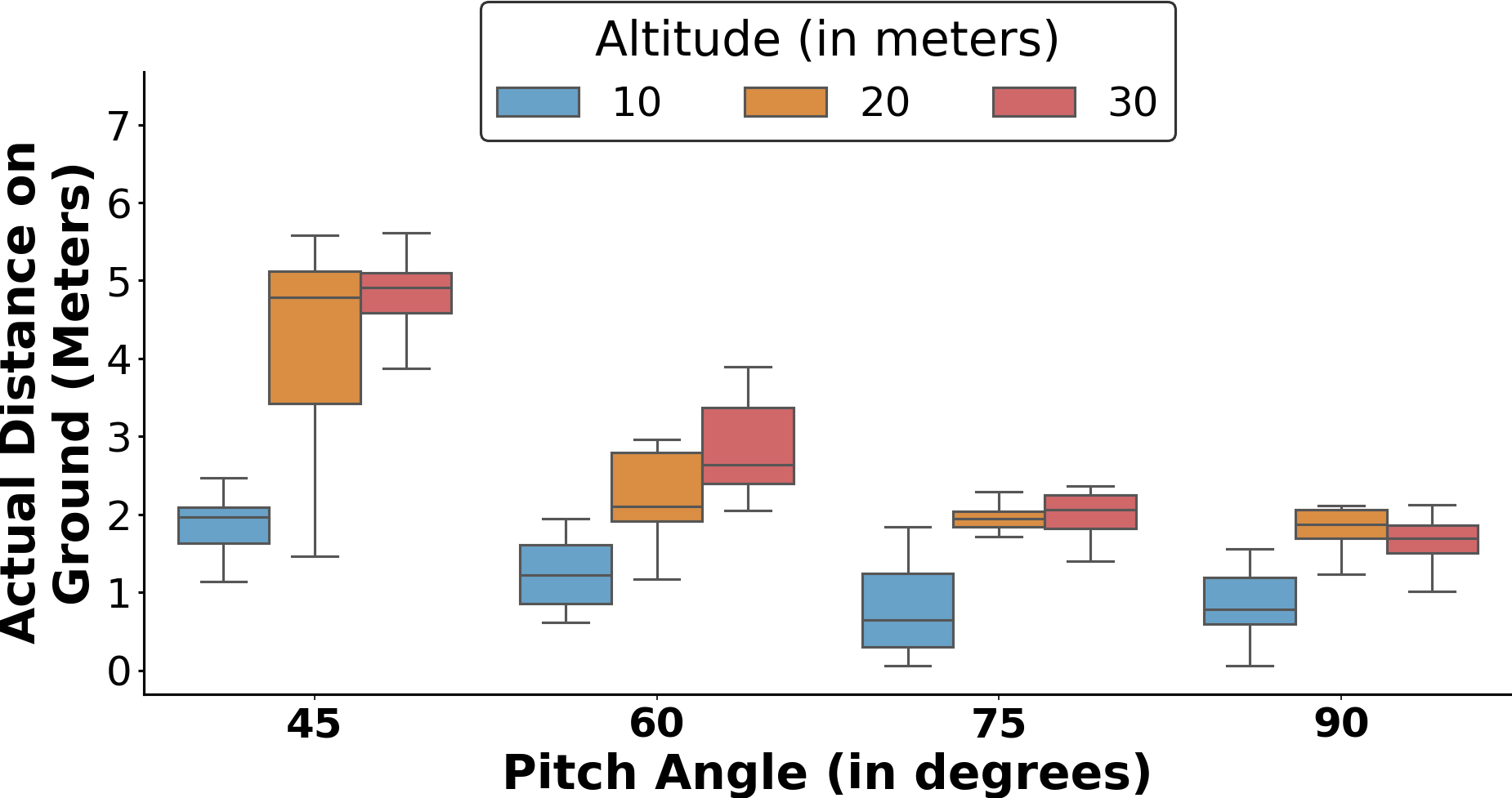}
    \caption{Geo Location Computation Accuracy}
    \label{fig:distance-pitch}
    \end{minipage}
    % \ankit{just need to increase the font size of scale and labels in the graph}
\end{figure}

The geodesic distance between the calculated location and the ground truth location of POI across different Altitudes and Pitch Angles are shown in Figure \ref{fig:distance-pitch}.

% \begin{figure}
%     \centering
%     \includegraphics[width=\columnwidth]{figures/analysis/Geolocation Computation/Altitude_violin.png}
%     \caption{Geo-Acc. - altitude}
%     \label{fig:enter-label}
% \end{figure}

% In these frames, we used OpenCV contour detection to find the center of the whiteboard. Each frame is manually reviewed to find any errors. 
% \ankit{why did you have to use opencv in simple geolocation accuracy calculation}
% Once we were sure about the center point of the whiteboard in each frame, we looked at the frame's metadata. This information was very important for our algorithm that calculates the geolocation, making sure that the location data we got was both accurate and relevant for each specific frame.
% \vspace{-12px}
Figure \ref{fig:distance-pitch} illustrates a decrease in horizontal accuracy (distance between computed and actual geolocation) with an increase in the altitude of the UAV. Specifically, the mean accuracy at altitudes of 10m, 20m, and 30m were recorded as 1.3 meters, 2.5 meters, and 2.9 meters, respectively. This pattern is largely due to the decreased spatial resolution in the imagery captured by the UAV at elevated altitudes, where each pixel represents a broader area of the ground. Thus, when trying to locate a POI in a high-altitude UAV's aerial video stream on the screen, even a small pixel deviation can significantly increase the geolocation computation error.
% \ankit{explain why accuracy decreased with increase in altitude}

Figure \ref{fig:distance-pitch} also shows that the accuracy of the computed geolocation decreases as the camera's pitch angle becomes more acute, meaning as the camera moves away from a position perpendicular to the ground. The highest accuracy is achieved when the camera is directly facing the ground (perpendicular). The mean accuracy at pitch angles of 45°, 60°, 75°, and 90° are 3.6m, 2.2m, 1.8m, and 1.5m, respectively. When the camera's pitch angle moves away from perpendicular, each pixel in the video covers a larger ground area, making it essential to account for depth information to compute accurate geolocations. However, our current geolocation method does not factor in depth, leading to increased errors as the camera pitch becomes more acute.
% \begin{figure}[ht]
%     \centering
%     \captionsetup{justification=centering}
%     \begin{minipage}{0.48\columnwidth}
%         \includegraphics[width=\columnwidth]{figures/analysis/contour_1.png}
%     \caption{Midpoint Detection}
%     \label{fig:contour_geo}
%     \end{minipage}%
%     \hfill
%     \begin{minipage}{0.48\columnwidth}
%         \includegraphics[width=\columnwidth]{figures/analysis/contour_2.png}
%     \caption{Pixel Distance}
%     \label{fig:contour_pixel}
%     \end{minipage}
% \end{figure}

% \begin{figure}
%     \centering
%     \includegraphics[width=\columnwidth]{figures/analysis/contour_marked.jpg}
%     \caption{Caption}
%     \label{fig:enter-label}
% \end{figure}

\subsection{AR Marker Visual Accuracy Analysis Method} We designed our method to measure the pixel distance between the center of the AR marker appearing on the aerial video frame and the actual center of the POI in the same frame. First, we hard-coded an AR marker at the POI's ground truth geolocation in the AR world space of the drone AR system. This ensures that the marker would automatically appear as an overlay on the aerial video whenever the UAV's camera view captures the corresponding geolocation space, as discussed in Section \ref{subsec:marker_augmentation}. Secondly, we employed the same contour detection image processing technique to identify the centers of both the AR marker and the POI visible in the frame. We subsequently calculated the distance between the two central points and analyzed the Euclidean distance between these midpoints. During our analysis, whenever the POI and the augmented marker overlap in the frame, we consider the pixel distance to be zero, suggesting an accurate representation of the POI from the user's perspective. From our collection of aerial videos, we selected the same 10 frames that we selected for analyzing the geolocation computation accuracy. 

% In each frame, two distinct white objects were present: one represented the augmented POI marker, and the other was the POI itself. After identifying the midpoints of these two white objects in each frame, we proceeded to calculate the Euclidean distance between these midpoints. 

% \begin{figure}[htbp]
%     \centering
%     \includegraphics[width=\columnwidth]{figures/analysis/Marker Augmentation/Altitude_violin.png}
%     \caption{Marker Augmentation Accuracy at different altitudes.}
%     \label{fig:pixelDistance-altitude}
% \end{figure}

% \begin{figure}[htbp]
%     \centering
%     \includegraphics[width=\columnwidth]{figures/analysis/Marker Augmentation/Pitch_violin.png}
%     \caption{Marker Augmentation Accuracy across different pitch angles. }
%     \label{fig:pixelDistance-pitch}
% \end{figure}

\begin{figure}[t]
    \centering
    \captionsetup{justification=centering}
    
    \begin{minipage}{0.86\columnwidth}
        \includegraphics[width=\columnwidth]{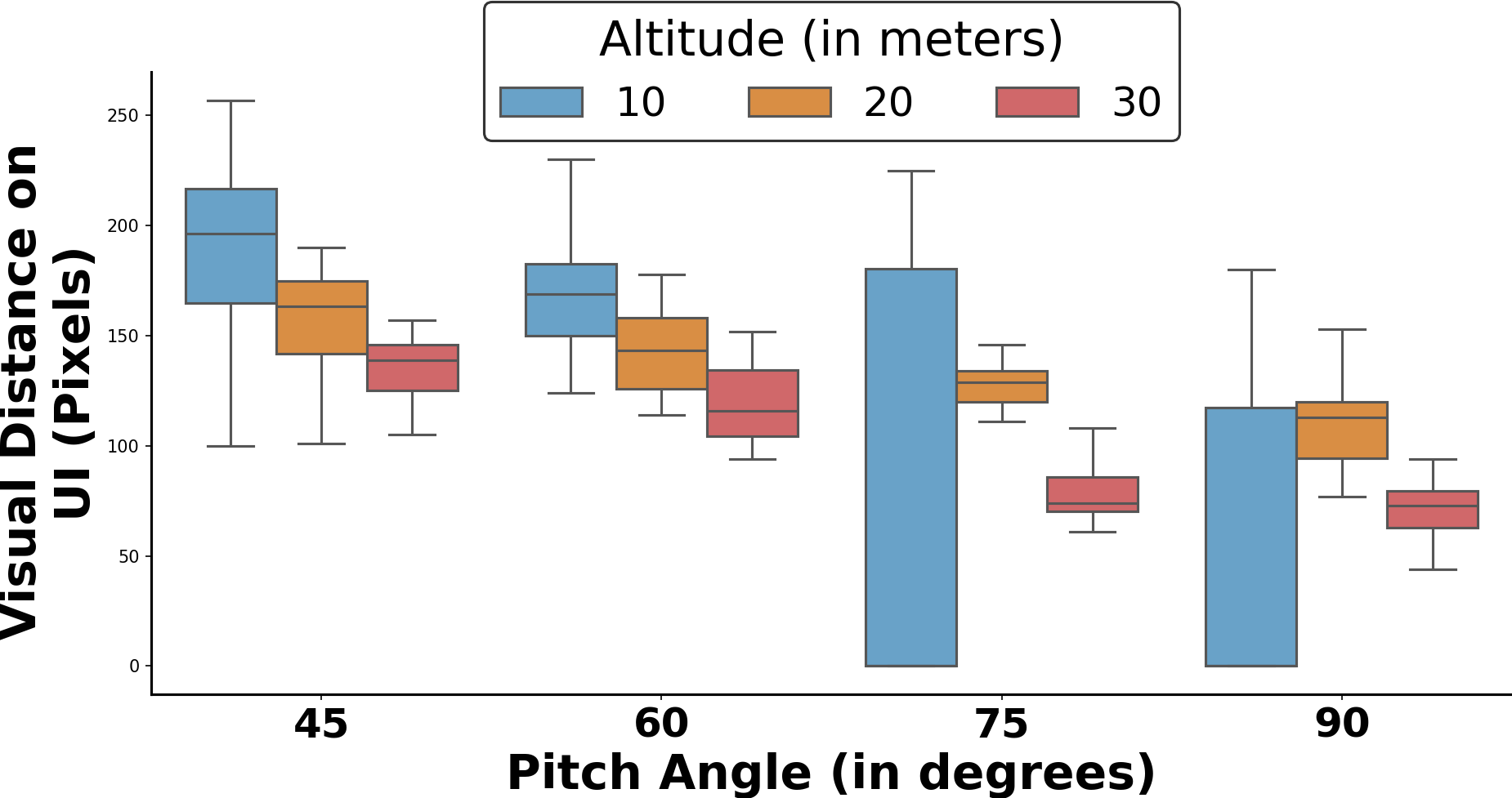}
        \caption{AR Marker Visual Accuracy}
        \label{fig:pixelDistance-pitch}
    \end{minipage}
    % \caption{Marker Augmentation Acc - Altitude and Pitch}
    % \label{fig:pixelDistance-alt-pitch}
\end{figure}

% \begin{figure}[htbp]
%     \centering
%     \includegraphics[width=\columnwidth]{figures/analysis/Marker Augmentation/Satellite_violin.png}
%     \caption{Marker Augmentation Accuracy across different satellite count. }
%     \label{fig:pixelDistance-satellite}
% \end{figure}
% \vspace{-12px}

\subsection{Analysis - AR Marker Visual Accuracy}
Figure \ref{fig:pixelDistance-pitch} shows the distribution of pixel distance between the augmented marker and POI across various altitudes. First, at an altitude of 10 meters, we notice a high variance in pixel distance. This variance can be attributed to two reasons. First, at such lower altitudes, since the AR camera is closer to the ground, the actual distance (in meters) between two points on the screen is high compared to higher UAV altitudes, meaning that even minor differences between the computed and actual geolocations of the POI result in a higher pixel distance between the AR marker and the center of the POI on the screen. Second, we considered a pixel distance of 0 if an AR marker overlaps partially or completely with the actual POI. As a result, in cases when the geolocations are calculated precisely at lower altitudes (refer Figure \ref{fig:distance-pitch}) the pixel distance metric remains zero. 

Figure \ref{fig:pixelDistance-pitch} also shows that as the camera pitch angle deviates from a perpendicular position to the ground (pitch angle approaches 45 degrees), the pixel distance increases. This increase occurs because the computational accuracy of the POI diminishes as the camera pitch angles deviate from this perpendicular position, causing the appearance of AR markers on the screen to shift away from the POI.

% whereas, as the pitch angle approaches a position perpendicular to the ground (90 degrees), the pixel distances decrease. S The average pixel distances for pitch angles of 45, 60, 75, and 90 degrees are 160, 142, 97, and 74, respectively.
% Average pixel distance decreased with increase in satellite count as shown in Figure \ref{fig:pixelDistance-satellite}.

% \section{}
\section{RQ2 - Field Study and  Analysis}

We deployed our \system in the real-world and conducted a field study to assess its performance in law enforcement activities and compared to traditional radio methods.

\begin{figure}[htbp]
    \centering
    \captionsetup{justification=centering}
    % Image 1
    \begin{minipage}{0.49\columnwidth}
        \includegraphics[width=\linewidth, scale=0.7]{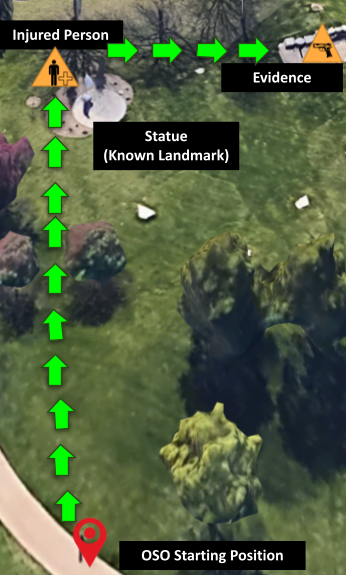}
        \caption{Search and rescue (known landmark)}
        \label{fig:sr-c1}
    \end{minipage}
    \hfill
    % Image 2
    \begin{minipage}{0.49\columnwidth}
        \includegraphics[width=\linewidth, scale=0.7]{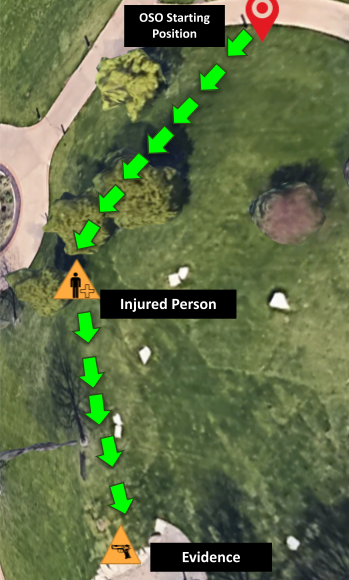}
        \caption{Search and rescue (unknown landmark)}
        \label{fig:sr-c2}
    \end{minipage}
\end{figure}

% \subsubsection{Search and Rescue}
\subsection{Field Setup}
We recreated a search and rescue mission common to law enforcement officers in a controlled outdoor environment. Here, participants, acting as OSOs and ROs, collaborated to locate a victim and key evidence, represented by six-foot-long posters and a lookalike gun, within a predefined area. The ROs were responsible for overseeing the area via aerial footage from a pre-configured Parrot ANAFI Quadcopter UAV\cite{ParrotAN2:online}, identifying key POIs like the victim and evidence, and communicating their geolocations to the OSOs, whose task was to navigate the area and locate these POIs.To further mitigate biases in the search and rescue task, two distinct conditions were established: SR-KnownLandmark and SR-UnknownLandmark. In SR-KnownLandmark, the victim's location was strategically placed near a unique statue (Refer to Figure \ref{fig:sr-c1}), providing a clear reference point for communication. Conversely, in SR-UnknownLandmark, the communication challenge was increased by positioning the victim behind one of many trees (Refer to Figure \ref{fig:sr-c2}). Additionally, to prevent the teams from relying on preconceived information, the position of the gun, representing key evidence, was varied between these two conditions. This approach ensures a more robust evaluation of the communication methods under different levels of complexity in the task environment.

\subsection{Participants}
In our initial field study, we engaged a total of 16 graduate students, organized into 8 teams comprising ROs and OSOs, with each task being performed by 8 teams across both conditions. The participant underwent thorough briefing and training to effectively perform the roles of ROs and OSOs. Tasks are assigned as follows: 

\noindent \textit{RO Tasks}: The RO's main task involved using live aerial footage to locate key POIs like the victim and evidence and then relay their geolocations to the OSO.

\noindent \textit{OSO Tasks}: The OSO's task was to navigate the area, based on information provided by the RO, and locate the POIs.

Each participant was compensated \$10 for their participation in a session that lasted approximately an hour. This initial field study allowed us to test the system in a real-world setting and prepared us for another study with police officers, which we plan to conduct in the near future.

\subsection{Study Design}
The study was structured into four phases. First, participants were briefed on \system and Human-UAV interaction research, covering technology use and study objectives. The second phase involved training with the communication tools, including both radios and the \systemx. The training session ensured that participants were familiar with and confident in using the RO-AR and OSO-AR systems before performing the actual study tasks. Following the training, the actual tasks were assigned to the teams, with specific roles designated to each member. To minimize biases, the initial communication system used for task completion, whether Radio or \systemx, was counterbalanced across the teams. Each team executed the assigned task under both conditions.  Further, to ensure robust and fair data collection, roles and field setups were switched before executing the task in the second condition. Finally, Upon completing the task in each condition, participants filled out a survey questionnaire.

\begin{figure}[htbp]
    \centering

        \includegraphics[width=0.85\columnwidth]{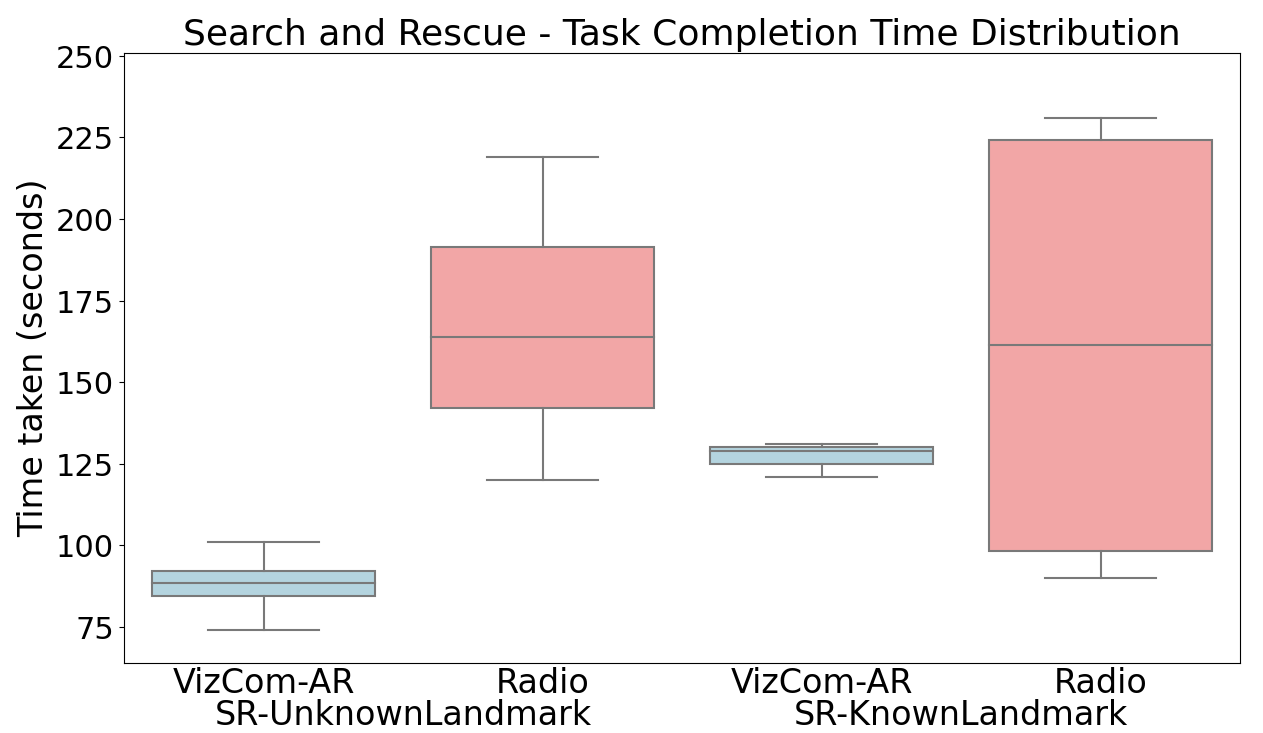}
        \caption{Time taken - \system Vs Radio}
        \label{fig:time-sr}
    
\end{figure}

\begin{figure}[htbp]
    \centering

        \includegraphics[width=0.85\columnwidth]{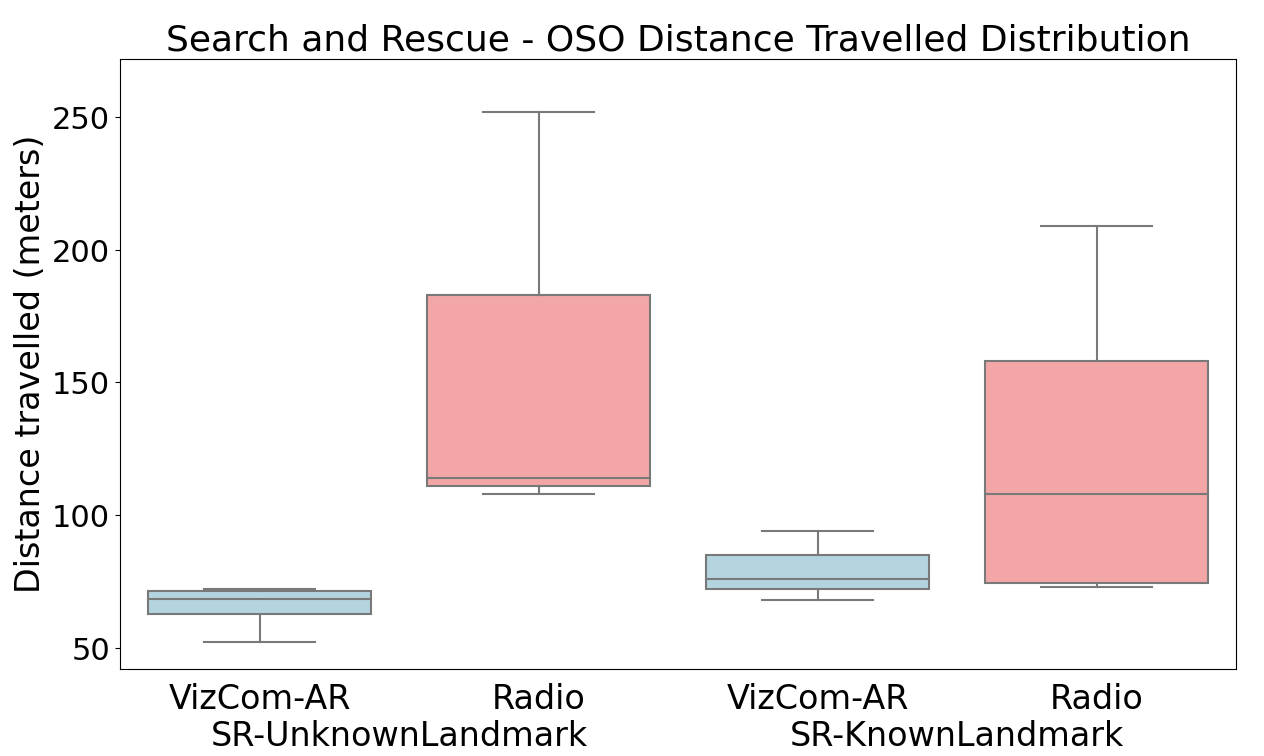}
        \caption{Distance traveled by OSO -  \system Vs Radio}
        \label{fig:distance-sr}

\end{figure}

\begin{figure*}[htbp]
    \centering
    \includegraphics[width=0.8\textwidth]{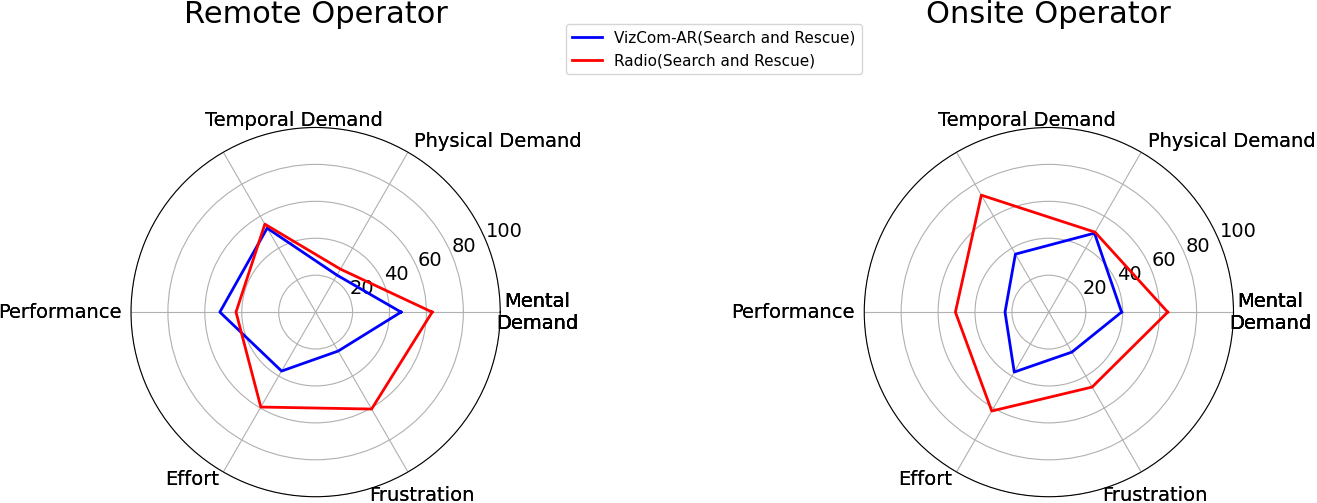}
    \caption{ NASA TLX scores for RO and OSO using both \system and Traditional radio communication}
    \label{fig:nasa-tlx}
\end{figure*}

\subsection{Data Collection} During the study, we collected the GPS positions of OSO throughout the study to investigate the total distance traveled by OSO to complete the task. Additionally, we used a stopwatch to record the total time each team took to complete the assigned task. We also collected responses to our survey questionnaire to specifically gauge both OSO's and RO's  Communication Quality.

\subsection{Team Performance}

Figure \ref{fig:time-sr} illustrates the distribution of time taken by teams to complete the search and rescue tasks using both Radios and \System. Overall, participants demonstrated a more consistent and effective performance when using \system as opposed to Radios. Teams using \system showed improved median completion times with reduced variations when compared to Radio-based communication across the two conditions (SR-UnknownLandmark and SR-KnownLandmark). In the case of SR-UnknownLandmark, teams faced challenges in completing the task while using Radio-based communication(median time of 164 seconds, IQR 191–142 seconds) whereas teams that performed the same task using \system showed significant improvement in task completion time(median time of 88 seconds, IQR 92–84 seconds).

Similarly, in the SR-KnownLandmark condition, teams using \system were more consistent, finishing the task in approximately 129 seconds (IQR 125-130 seconds). When using Radio, time varied; some teams used the landmark as the anchor to precisely communicate the location of the victim in the scene. This led to a wider spread of completion times, with a median of 161 seconds and an IQR of 98-224 seconds. Moreover, the substantial variations in time (as indicated by the wider inter-quartile range) reveal that individual differences in processing and communicating information over the radio played a role in the overall performance of the team. The total distance traveled by OSOs further reinforces these results. Utilizing Radio-based communication, OSOs were again able to minimize the distance traveled when the victim was near a unique landmark but faced challenges in scenarios with ambiguous environmental cues. With the use of \system system, the distance traveled by OSO remained consistently shorter across both conditions, as shown in Figure \ref{fig:distance-sr}.

\subsection{Team Workload}

The NASA Task Load Index (TLX) results, as shown in Figure \ref{fig:nasa-tlx}, provide evidence of \system efficacy in alleviating the workload for both RO and OSO. Both RO and OSO experienced lower mental demand (score reduced from 63 to 46), frustration (score reduced from 60 to 24), and effort (score reduced from 59 to 36) when compared to using radio communication.  On the other hand, while we expected improved perceived performance for both OSO and RO, participants felt they performed better with Radio than with \System. This suggests that while \system helped reduce the cognitive load for communicating geolocations and enhanced the RO's experience, further research is needed to explore how training both RO and OSO can improve the effective use of Augmented Reality in their workflow for better performance.

\begin{figure}[htbp]
    \centering
    \includegraphics[width=0.45\textwidth]{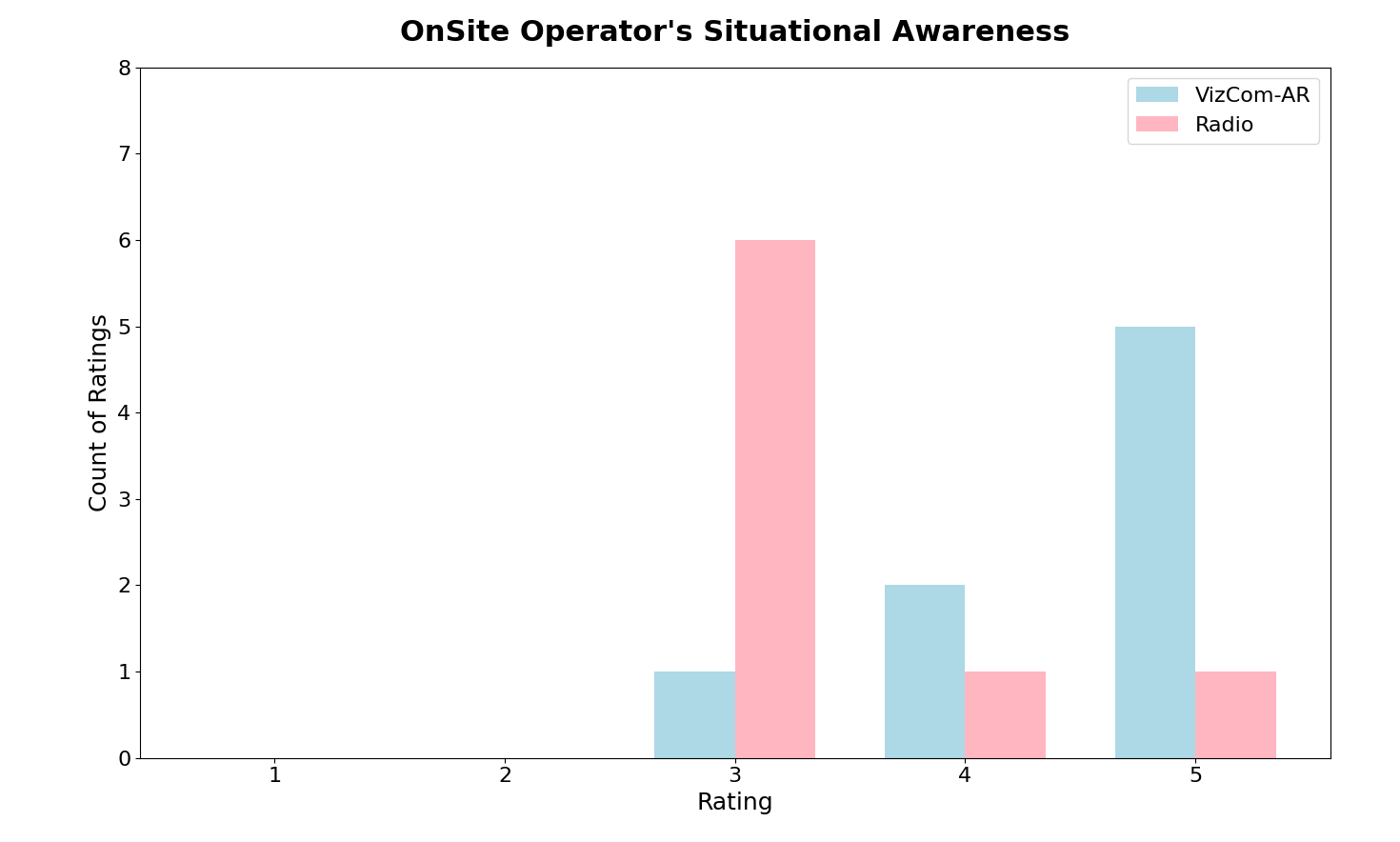}
        \caption{OnSite operator's situational awareness }
        \label{fig:situational-awareness-oso}
\end{figure}
\begin{figure}[htbp]
    \centering
    \includegraphics[width=0.45\textwidth]{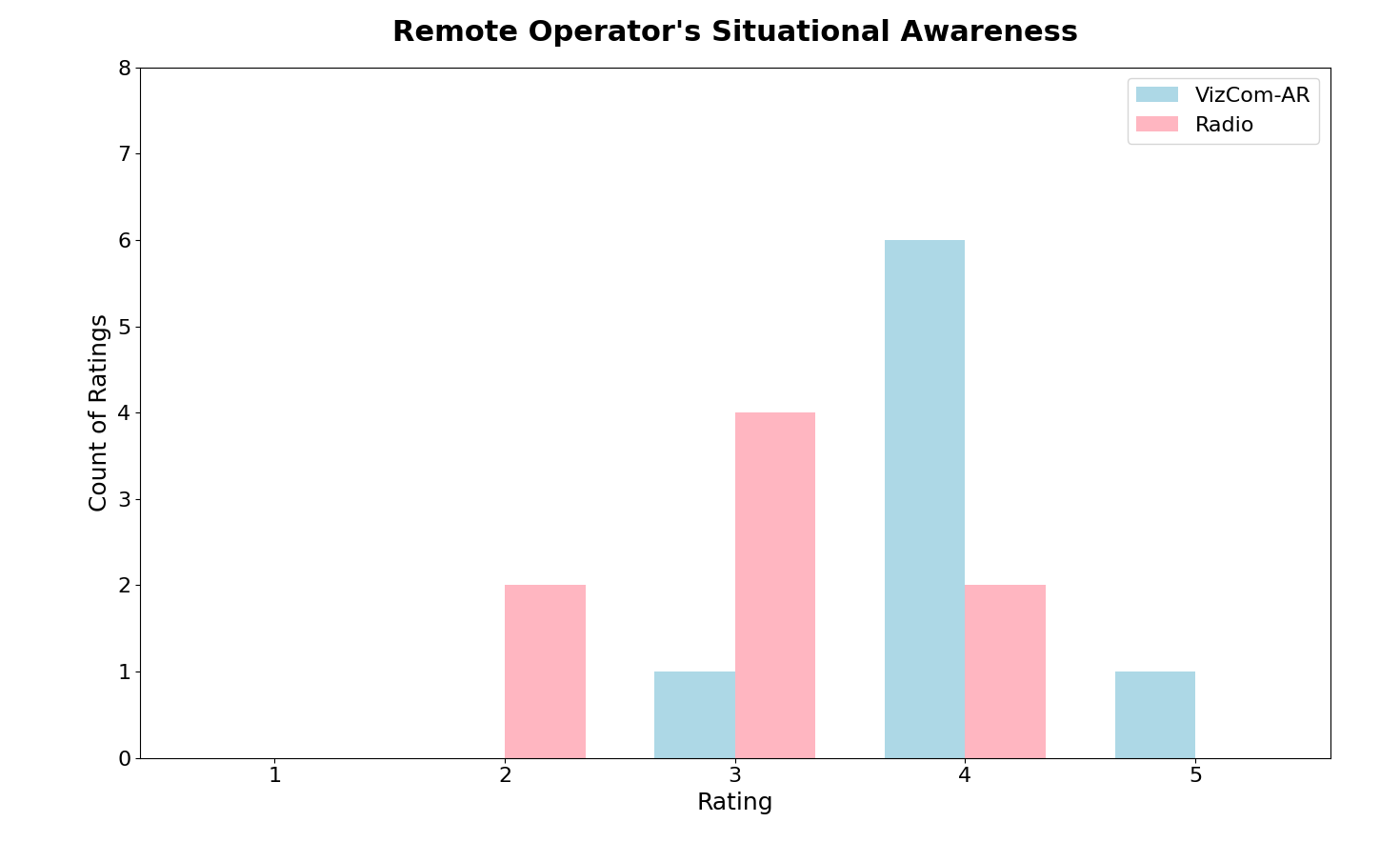}
        \caption{Remote operator's situational awareness}
        \label{fig:situational-awareness-ro}
\end{figure}

\subsection{Situational Awareness}
To gauge the situational awareness of both ROs and OSOs, we used a mixed-method approach that incorporated both a 5-point Likert scale and open-ended questions. The quantitative SA scores, illustrated in Figure \ref{fig:situational-awareness-oso} and Figure \ref{fig:situational-awareness-ro}, show a substantial enhancement in SA when using \system as opposed to traditional Radio-based Communication across all conditions. Notably, OSOs experienced a marked increase in SA when using \system. Feedback from the open-ended questions complements these findings. OSOs specifically noted that \system’s visual cues made the identification of POIs straightforward, reducing the cognitive load, improving awareness, and aiding in task completion.

\section{Conclusion and Future Work}
\label{sec:conclusion}
We presented \system that allows police officers or rescue mission personnel, including ROs who monitor the scene via aerial video streams, and OSOs who work at the scene, to communicate crucial geo-location information in complex environments precisely through Augmented Reality. Police officers validated the design and usability of the system in a preliminary focus-group session. We evaluated our system's effectiveness under various UAV flight parameters, including changes in altitude and pitch angle. Furthermore, through a user study, we discovered advantages of using our system over traditional radio communication methods in emergency scenarios. Participants using our system completed tasks with greater efficiency and reported significantly improved situational awareness. This work opens up new research avenues, especially for the study of augmented reality technology to design complementary tools for data visualization and communication during UAV-driven search-and-rescue missions. The findings and lessons learned from this study provide a foundation for future advancements in the domain of AR-enhanced UAV-driven emergency response. In future, we aim to scale \system to support multi-UAV missions. This enhancement would allow annotations from one aerial video stream to be automatically shared with other ROs monitoring different UAVs.

% We presented \system that allows emergency response personnel, including ROs who monitor the scene via aerial video streams, and OSOs who work at the scene, to communicate crucial information in complex environments precisely through Augmented Reality. Our approach to calculate the location of POI using the pinhole camera model and augmenting marker in the UAV video stream was rigorously evaluated across various flight characteristics, including altitude and pitch angle. The user study conducted as part of this research highlights the significant advantages of our AR framework over traditional radio communication methods in emergency scenarios. Participants who utilized our system not only completed tasks more efficiently but also experienced enhanced communication quality. This study provides a valuable foundation for future advancements in the field of Augmented Reality Systems for UAVs.
\section{Acknowledgement}
We thank the Saint Louis University Research Institute for funding this work and Dr. Jane Cleland-Huang for moderating the focus group session and initial discussions during the conception of this project.

\bibliography{AR}
\bibliographystyle{abbrv}
\end{document}